# CAN MACHINE LEARNING TOOLS SUPPORT THE IDENTIFICATION OF SUSTAINABLE DESIGN LEADS FROM PRODUCT REVIEWS? OPPORTUNITIES AND CHALLENGES

**Michael Saidani[1], Harrison Kim**
Department of Industrial and Enterprise
Systems Engineering, University of Illinois at
Urbana-Champaign, Illinois, USA

**Bernard Yannou**
Laboratoire Genie Industriel, CentraleSupélec,
Université Paris Saclay,
Gif-sur-Yvette, France

## ABSTRACT

*The increasing number of product reviews posted online is a gold mine for designers to know better about the products they develop, by capturing the voice of customers, and to improve these products accordingly. In the meantime, product design and development have an essential role in creating a more sustainable future. With the recent advance of artificial intelligence techniques in the field of natural language processing, this research aims to develop an integrated machine learning solution to obtain sustainable design insights from online product reviews automatically. In this paper, the opportunities and challenges offered by existing frameworks – including Python libraries, packages, as well as state-of-the-art algorithms like BERT – are discussed, illustrated, and positioned along an ad hoc machine learning process. This contribution discusses the opportunities to reach and the challenges to address for building a machine learning pipeline, in order to get insights from product reviews to design more sustainable products, including the five following stages, from the identification of sustainability-related reviews to the interpretation of sustainable design leads: data collection, data formatting, model training, model evaluation, and model deployment. Examples of sustainable design insights that can be produced out of product review mining and processing are given. Finally, promising lines for future research in the field are provided, including case studies putting in parallel standard products with their sustainable alternatives, to compare the features valued by customers and to generate in fine relevant sustainable design leads.*

Keywords: sustainable design, data-driven design, machine learning, natural language processing, product reviews.

## 1. INTRODUCTION

### 1.1 Context and motivations

A large number of product reviews are posted online every day. This is both (i) a convenient way for customers to make their voice heard, and (ii) an opportunity for designers to improve the features of their products on this basis [1, 2]. In parallel, artificial intelligence (AI) is transforming and improving processes in many areas and creates new opportunities and challenges for designers. Indeed, the recent advance of information technology (IT) tools (e.g., web scraping) and machine learning (ML) techniques (e.g., natural language processing) enable researchers and industrialists to extract, proceed, and analyze broad datasets of online reviews. In product design and development, companies can now have the ability to acquire a large quantity of information in a short time and at a low cost, thanks to the support of recently developed IT and ML tools.

Yet, in their reflection on how the practice of product development might be developing over the next 20 years, Isaksson and Eckert (2020) stated that recent technological innovations, such as machine learning, have been around since the 1990s but are reaching industrial practices in product development at a slow pace, because most industrialists do not have awareness and know-how to capitalize on them [3]. Likewise, in their research roadmap for the next ten years that could revolutionize how sustainable design is practiced (i.e., to better build sustainability into the way all products and services are designed), Faludi et al. (2020) recognized that the development of sustainable design methods and tools must take into account the changes and needs of an industry that will evolve towards more digitalized product development practices [4]. "Understanding users and usage", and "integrating data and analytics" are also two of the main challenges identified by Kim et al. (2020) for the future of ecodesign practices [5].

---
[1] Contact author: msaidani@illinois.edu

In this line, the present study starts investigating and discussing how machine learning techniques and tools could be deployed to advance the sustainable design and use of products. Notably, it aims at developing and training a machine learning process, combining suitable automated natural language processing (NLP) techniques, to get sustainable insights from online product reviews. The umbrella expression "sustainable design insights" encompasses here any information that could be used to design more sustainable products all along their life cycles [4, 6]. An illustrative list of such sustainable insights, learnings, or leads, is given in the results section of the paper. As such, this piece of research contributes thus to developing theoretical and practical knowledge in data- and AI-centric sustainable design methods and tools, by discussing and developing novel approaches to assist designers and organizations in implementing sustainable design models and ultimately reducing the environmental impact of products and services.

**1.2 Current gaps and research objectives**

As of now, few research papers have tried to linking product sustainability features and sentiment analysis from online reviews, and existing works present several shortcomings, notably in terms of automation and sustainable design implication [2]. Moreover, Kwok et al. (2020) noticed a gap between how product sustainability information is perceived by customers and how this impacts the purchasing behavior of customers [7]. Also, a recent study performing sentiment analysis on sustainability reporting [8] has highlighted a key issue: different sentiment analysis tools, models, or algorithms (which are described in sub-section 3.2) can produce different outcomes – i.e., polarities (positive, negative, or neutral) – while using the same dataset.

Meanwhile, during a webinar of the Design Society hosted by the Sustainable Design Special Interest Group, Borgianni (2020) started to question how users perceive the products' sustainable performance [9]. On this basis, there are further leading questions that are worth investigating: Are people aware of sustainable advantages and value ecodesign products accordingly? How much can we learn from customers' product reviews to design more sustainable products? What design learnings could be elicited from online product review? How do we develop greener products based on these insights? Mining online product reviews for sustainable product development is the overall objective of this research work. More specifically, this project aims to question and discuss the potential of ML and NLP techniques to: (i) improve effectively the environmental friendliness of products (through new AI-based approach using ML tools to design eco-improved products), (ii) better know the usage behavior and attitude of customers towards eco-products, so that they are less likely to be used in a non-sustainable way.

The present paper starts analyzing how ML tools could be deployed to help generating new insights from online product reviews to design more sustainable products and services. Notably, the opportunities and challenges for each step of the ML process are discussed, as illustrated in Figure 1. Note that a complementary paper submitted as well as this conference provides concrete examples, extracted manually, on sustainable design leads that can be identified and interpreted from online product reviews [10].

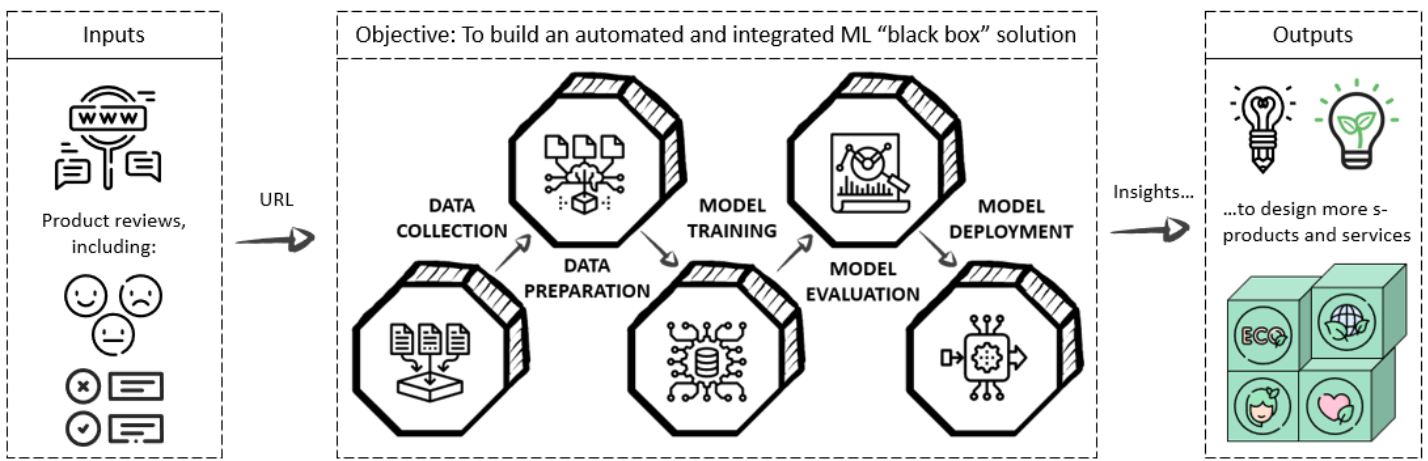

**FIGURE 1:** WORKFLOW OF THE MACHINE LEARNING PROBLEM TO SOLVE

## 2. RESEARCH APPROACH

The overall research methodology of this project is based on two complementary parts, as it follows: the first part is to review and examine the available studies and tools for product reviews mining, to discuss the strengths and opportunities offered by existing works and frameworks, as well as their limitations to be deployed in our context, i.e., to lead to new insights in the design and development of sustainable products; on this basis, the second part aims to develop an integrated solution (an easy-to-deploy "black box", as illustrated in Figure 1) to solve this ML problem, by (a) combining, adjusting, and/or training the suitable models, and (b) testing and fine-tuning it on case studies.

The scope of the present paper not only covers the first part, i.e., after defining, making sense, and illustrating the ML problem to solve, we dive into the ML process (and its associated steps) through literature, code, and tools surveys and analyses. But also, it brings out and discusses key elements to properly conduct the second part of this project, as illustrated in Figure 2. First, a screening of both the academic literature using Scopus and Google Scholar databases, and the grey literature through Google search has been done with a combination of the following keywords: product, design, eco-design, sustainability, sustainable, feature, online review, sentiment analysis, opinion mining, text mining, artificial intelligence, machine learning, and natural language processing. Note that text mining, opinion mining, and sentiment analysis are often used interchangeably to refer to a group of NLP techniques. As aforementioned, this literature survey has shown a lack of research at the intersection of sustainable product design and online review mining.

With this background, this study then investigates and questions how can existing NLP techniques be borrowed, combined, and/or modified to fill this gap. Notably, most acknowledged and state-of-the-art NLP techniques are discussed and positioned on a generic machine learning process. Finally, the implications for designing and developing more sustainable products are examined accordingly.

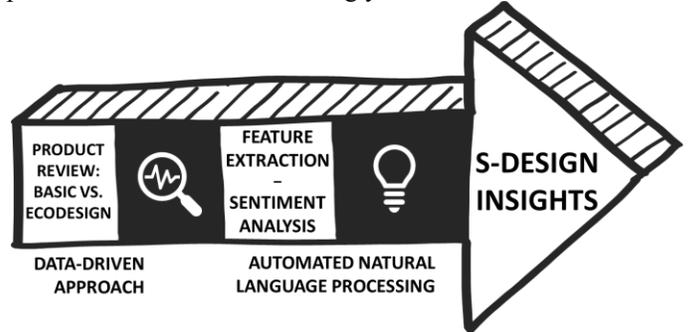

**FIGURE 2:** ILLUSTRATION OF THE VISION AND RESEARCH APPROACH: FROM COMPARATIVE PRODUCT REVIEWS TO SUSTAINABLE DESIGN (S-DESIGN) INSIGHTS

## 3. RESULTS AND DISCUSSION: BETWEEN CHALLENGES AND OPPORTUNITIES

In this section, a breakdown analysis of the generic machine learning (ML) process for natural language processing (NLP) is performed through the lens of generating leads for sustainable design. Each step of this process, as displayed in the flowchart of Figure 3, is analyzed. Notably, state-of-the-art ML tools and NLP algorithms are examined (from sub-section 3.1 to 3.5) in the light of their relevance, and limitations, in contributing to producing useful sustainable design insights.

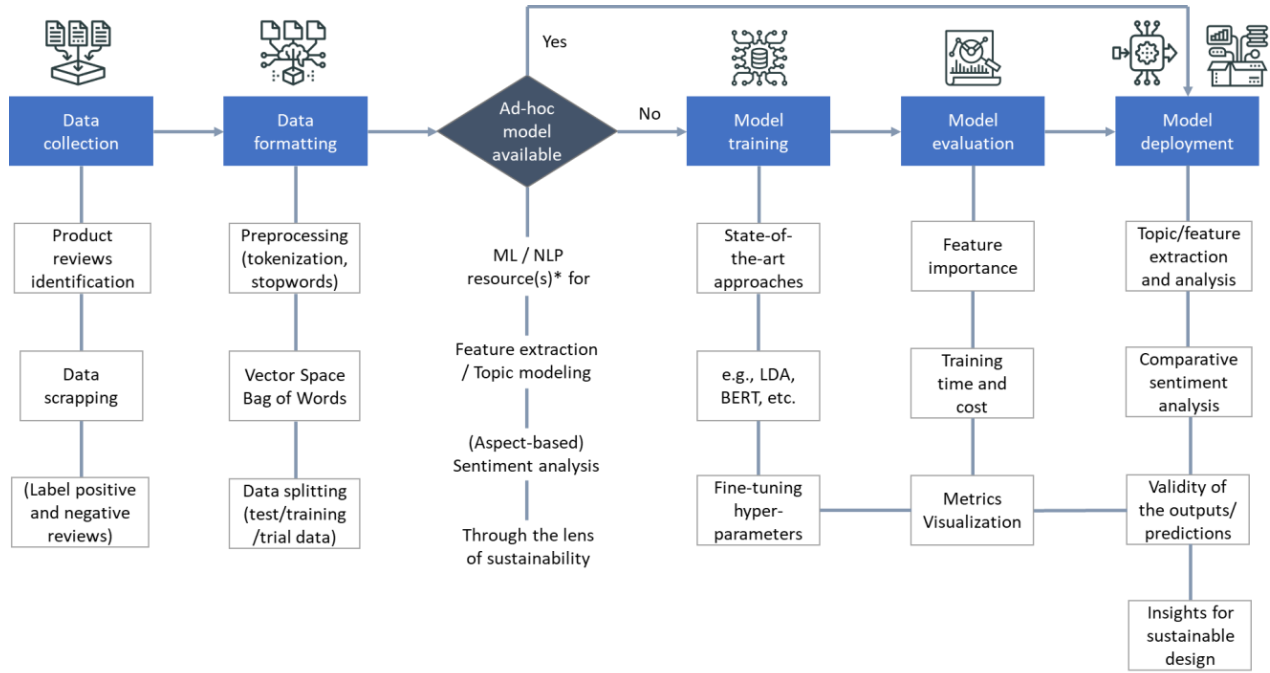

**FIGURE 3:** MACHINE LEARNING STEPWISE PROCESS FOR MINING PRODUCT REVIEWS
(*Resources = existing methods, ready-to-use models, pre-trained models, open-access algorithms and libraries)

## 3.1 Data collection: scraping sustainability-related product reviews

Online product reviews are becoming a valuable resource for designers in order to complete the limitations (time-consuming, substantial research costs, low samples of customer feedback) of traditional approaches to understanding customer perceptions, such as surveys, interviews, and focus groups [2, 11]. According to Isaksson and Eckert (2020), systemic and automated data collection will soon be an integrated part of the product development process [3].

On the one hand, there is a wealth of Python libraries and packages available, facilitating automated data collection from online product reviews. Notably, numerous product review scrapers accessible both for developers and designers, e.g., using either Python notebooks or web browser extension to extract the reviews directly in a "csv" Excel format. For instance, Beautiful Soup is a Python library that eases the process of scraping information from web pages, for developers or users that have at least a basic knowledge of Python code and functions. There are also many free and easy-to-use software or web-based versions for non-developers, such as webscraper.io, that are useful for designers, engineers, industrialists, and researchers who are lacking programming skills. Though, such tools have low robustness, i.e., if the online platform changes the layout or display format of the reviews, the scrapping tool might become obsolete and irrelevant. Note also that some online shopping (e-commerce) websites sometimes may block product review scraping when it is overused.

On the other hand, while an increasing number of product reviews is posted online, a relatively low number of products labeled as "sustainable" or "eco-friendly" is available on major e-commerce websites, such as Amazon [11]. For example, a small number of eco-bikes is sold on this platform, and the modular smartphone "Fairphone" is not available. Also, the popularity of these eco-products (quantified hereafter by the number of global ratings and written reviews) is currently significantly inferior to their standard alternatives. Both to further illustrate this point and to look for relevant case studies comparing the reviews from a conventional product to its sustainable alternative, a keyword-based search on amazon.com has been performed using the following queries: eco-X, eco-friendly X, green X, sustainable X, bio-based X, renewed X (where X represents here a specific product previously identified as potentially relevant, or the auto-suggested item given by the e-commerce platform). For instance, it has been found (as of mid-November 2020) that the standard top-reviewed (in terms of the number of global reviews available) wireless keyboard and mouse has a total of 3,023 global ratings with 1,673 global reviews. While its "green" top-reviewed alternative of the same category (i.e., having a similar size and main function) has a total of 153 global ratings with 112 global reviews.

The same story can be told in several product categories when comparing the number one standard top-reviewed product and its eco-friendly top-reviewed alternative, such as: a razor system made with recycled ocean plastic, a reusable metal razor, an eco-flow shower head, a solar-powered portable speaker, or an energy-saving LED TV. Note that for the panel of products listed here, the eco-friendly alternative product is on average slightly more expensive. In the meantime, one can argue that the current momentum on climate change, environmental awareness, and circular economy, would foster and catalyze the attractiveness for sustainable products. Notably, the apparition of product eco-labels on e-commerce platforms could be a support in that direction. The Amazon Climate Pledge Friendly actually recognizes products with improvements in at least one aspect of sustainability and that are certified by one of the sustainability certifications (e.g., Cradle to Cradle, Recycled Claim Standard, or Compact by Design). For example, Compact by Design products remove excess air and water, which reduces the carbon footprint of shipping and packaging. Also, there is an increasing number of renewed smartphones or laptops available on e-commerce platforms, but it remains to quantify if their overall impact is in favor of sustainability from a lifecycle perspective, in order to identify and filter out misleading or "fake" sustainable claims by marketers. Examples of reviews mentioning sustainability-related aspects – both on environmentally certified and standard products – are further analyzed and discussed in a complementary paper submitted at this conference [10].

## 3.2 Data formatting and pre-processing

As online reviews are unstructured and often written in different forms (e.g., fully written sentences, bullet lists, couple of keywords, etc.) for the same product [2], the reviews need to be properly formatted to be understood by ML models and NLP algorithms. Here, many ready-to-use and/or easy-to-adapt codes (e.g., using Python 3 and its libraries) can be deployed for automatized data cleaning and text processing, including tokenization, removing stop words, and document term matrix generation [12]. While this paper focuses on Python libraries and framework (because of the significant number of available studies and templates written in Python [13]), note that the data collection, data formatting, and model training phases could also be performed in other programming languages such as R or C.

On the other side, a key challenge is to filter out fake positive, e.g., between a truly sustainability-related statement and a greenwashing-related comment, and fake negative reviews, e.g., to discredit a new competitive eco-friendly product. Online data analytics tools, such as fakespot.com, can be used to identify fake reviews from authentic ones [2]. Interestingly, according to El Dehaibi et al. (2019), it can be assumed that the number of non-authentic reviews that contain sustainability aspects is small, knowing that fake reviews mostly have generic content [2]. Furthermore, the identification and interpretation of irony and sarcasm is another challenge when mining product reviews in general, that remains valid in the present case.

A last challenge here is that the context in which the reviews are posted is unknown to the designer most of the time. In this line, Hou et al. (2019b) proposed a framework to understanding how and in what condition customers use their products, in order to better draw innovation leads from online product reviews to support decision making during product design and redesign [14]. In addition, Suryadi and Kim (2019) recently developed a process to automatically identify product usage contexts from online customer reviews [15].

### 3.3 Ad-hoc classifier available or model training

Many existing libraries and packages are available (e.g., embedded in Python notebook) to perform sentiment analysis, topic modeling, and feature extraction on product reviews. Yet, to the best of our research, no library or dictionary available has been trained explicitly using datasets of sustainability-related topics, features, or wording. For instance, El Dehaibi et al. (2019) manually modeled (using annotations) these perceptions of product sustainability features using ML techniques to determine which of these features are associated with positive and negative sentiments [2]. However, their manual annotation process is a time-consuming procedure, including four consecutive steps performed by a human: (i) read product reviews, (ii) highlight texts that mention sustainability aspects, (iii) identify product features from highlighted texts, and (iv) evaluate emotions in highlighted text.

With this background, it becomes interesting to question first how existing frameworks perform through the lens of product sustainability features identification and modeling (e.g., for aspect-based sentiment analysis). Then, it is key to investigate how different techniques, tools, and algorithms can be combined or fine-tuned in order to answer our research question, i.e., how to generate insights for sustainable product design from online reviews. This can also contribute to the vision of Isaksson and Eckert (2020) on the future of product design and development, stating that "tools and techniques will need to be developed to aid engineers in analyzing data in a way that can improve product quality and inform the design of similar systems" [3].

In natural language processing (NLP), sentiment analysis (SA) is the field of study that analyzes people's opinions, sentiments, evaluations, attitudes, and emotions from written language. SA is one of the most active research areas in NLP and is also widely studied in data mining [16-19]. It is a text analysis technique that detects polarity within texts, whether a whole document, paragraph, or sentence. The polarity score of a sentence usually ranges between -1 and 1, indicating sentiment as negative to neutral to positive. There are many prominent open-source Python packages and libraries (such as TextBlob, Standford Sentiment Treebank, Stanford CoreNLP, which are well-known NLP libraries in Python) [20], as well as web-based tools [21] in place to perform SA on product reviews. Yet, when dealing with complex reviews, different SA models can provide different outcomes for the same dataset [8]. Most of these models are trained on different datasets (e.g., using movie or restaurant reviews) that are not appropriate for comparing on a sound basis the reviews of a standard product and its eco-friendly alternative.

One relevant solution here is to train a new ad-hoc machine learning (ML) model that will understand the context better. For instance, KPMG (2019) built a new ML model based on the pre-trained BERT NLP model to perform sound sentiment analysis on sustainability reports [8]. While developing a new ML model can be a challenging and time-consuming task, using pre-trained NLP models, such as BERT for specific fine-tuning of NLP, appears as a convenient and timely solution here. BERT, which stands for Bidirectional Encoder Representation from Transformers, is a pre-trained language representation model developed by Google researchers to perform and solve a variety of state-of-the-art NLP tasks, like text classification, translation, or summarization [22]. For classification tasks such as sentiment analysis, one can fine-tune the model by adding a layer for classification, that as the first token output of the BERT base model as input, to train and deploy a fine-tuned BERT model for aspect-based sentiment analysis [23] on product review comparison with the aim, e.g., to better know what sustainability features are valued by customers. Here, aspect-based sentiment analysis, also known as feature-based sentiment analysis, appears to be a suitable technique to identify fine-grained opinion polarity towards a specific aspect or product feature.

In this line, having a performant and adequate model for product feature extraction is essential, in order only to assign a polarity score to the full review through sentiment analysis, but also to gain further and granular-level insights on which features are considered as positive or negative by the customers. Different acknowledged techniques exist for product features extraction, such as "Bag of Words" or "Term Frequency – Inverse Document Frequency" [12], for which many notebooks written in Python can be found online. Topic modeling algorithms, such as Latent Dirichlet Allocation (LDA), can also be a relevant solution to uncover the hidden thematic structure in a collection of product reviews [11. 24, 25]. For instance, Wang et al. (2018) extracted key topics of online reviews for two competitive products using LDA [11]. While the topic heterogeneity analysis demonstrates the unique topics of the two products, and allows to extract the competitive superiorities and weaknesses of both products, the topic labels had to be named manually.

To use such a topic modeling technique, a document-term matrix (see sub-section 3.2) is needed in input, as well as the number of topics to be picked by the algorithm. Once the topic modeling technique is applied, the human role is to interpret the results and decide whether or not the mix of words in each topic makes sense [12]. If not, one can try changing up the number of topics, the model parameters, or even try a different model. For information, El Dehaibi et al. (2019) came up with a first list of sustainable-related topics to look for in product reviews, e.g., for environmental aspects, the product features generated from topic modeling revolved around durability, material use, air emissions, and energy and water consumption [2]. Examples of sustainable design learnings and insights that could be elicited from this processing of product reviews are given in sub-section 3.5. Eventually, while there are numerous publicly available datasets and Python libraries (such as StanfordNLP) to perform conventional topic modeling and feature-based sentiment analysis, the use of newly developed NLP deep learning tools appears as a promising line of research to train better models [26].

### 3.4 Model evaluation

When developing and training a new classification model, e.g., an aspect-based sentiment analysis model fine-tuned for evaluating the customers' perceptions of sustainable features, it is of the utmost importance to quantify its performance (in comparison with other models) to validate or improving it. On the one hand, four main – easy-to-compute, and easy-to-implement – metrics to evaluate the performance of a machine learning model are well-acknowledged by the community, namely: precision, recall, F-measure, and accuracy metrics. One can refer to Haque et al. (2018) for the definitions and formulas associated to these measures [12]. On the other hand, as there is no agreed definition of common attributes for "sustainable products", nor an exhaustive list for such attributes, it remains challenging to build an ad-hoc and robust evaluation system. In this line, to validate the fact that some features contribute to the sustainability of a product, it could become interesting to link this machine learning approach with life cycle analysis [27] to quantify and compare the environmental footprint of a product claiming to be "eco-friendly" to its standard version.

### 3.5 Model deployment to identify sustainable design leads

Hou et al. (2019c) made a short review and summary of how online review analysis (using product feature and opinion-based models) can be deployed for product design [28]. While no method was purely proposed in the published literature for automatically finding eco-improvement leads from product reviews, product review mining has already been used for: design trends monitoring, building market strategies, product longevity prediction, discovering product defects, and construction of product improvement strategy. Here are some examples of main sustainable design learnings or insights that could be elicited from mining and comparing online product reviews (non-exhaustive list):

- Positioning and importance of sustainability-oriented features or properties among critical product features or functions (e.g., mapped along the Kano model [29]);
- Customers' preferences to prioritize sustainability product/feature information [7];
- Positive/negative perceptions of sustainable features (e.g., use of plastic, unsafe to a user, like "glass cracking") [2];
- Non-sustainable use cases or patterns of a product supposed to be sustainable (resulting in a wasted potential) [9];
- Automated detection of defects on consumer products [30];
- Wear and tear, or an early failure, of specific components or parts, altering the proper functioning and lifespan of products.

In a related paper submitted at this conference [10], further concrete examples on how reviews can provide designers to come up with more sustainable products are given, including extracts of reviews mentioning sustainability-related aspects on three technical and electronic products (laptop, printer, and cable). Furthermore, the linkages between product features, affordances, and customers' emotions, perceptions, or usage contexts are more accessible than ever to guide new product development. For instance, Hou et al. (2019c) analyzed the correlation between online reviews and product innovation to propose relevant design innovations for the next generation of a line of products [28]. One can imagine a similar or hybrid approach to generate relevant leads for eco-innovation. Eventually, comparing the reviews of a standard product to his greener alternative could lead to other findings or insights, e.g., for marketing purposes: How (differently) eco-friendly products are perceived by consumers (e.g., more expensive, better quality, more sensible)? How do sustainability features rank among other product attributes classified into the five categories of the Kano model for product development and customer satisfaction: must-be attributes, performance attributes, attractive attributes, indifferent attributes, and reverse attributes [29]?

Eventually, here are some further examples of how machine learning models can be deployed to better explore and exploit the (sustainable) design space. Kwok et al. (2020) started discussing, through a specific case study, the potential of (i) identifying prioritized sustainability attributes using sustainability design space, and (ii) applying machine learning to model customer preferences [7]. In this direction, applying machine learning in the aerospace industry, Bertoni et al. (2020) analyzed the potential for automating the value and sustainability assessment in design space exploration [31]. In a more advanced stage, Hein and Condat (2018) questioned whether machines could design, i.e., to come up with creative solutions to problems and build tools and artifacts across a wide range of domains [32]. These authors proposed a Gödel machine framework to generate novel design concepts, but a proof of concept implementation remains to be performed. In the present context, examining how these frameworks could be used or modified using product reviews in input could be relevant to come up with more insightful outcomes.

In all, based on this state-of-the-art literature review on NLP techniques, Table 1 provides a list of the opportunities to reach and the challenges to address for developing an ML process in order to get insights from product reviews to design more sustainable products.

**TABLE 1:** SYNTHESIS OF THE OPPORTUNITIES AND CHALLENGES FOR EACH STEP OF THE MACHINE LEARNING PROCESS

| ML and NLP processes and tools to design more sustainable products - Application to online review mining | Opportunities | Challenges |
|---|---|---|
| Data collection | - Increasing number of product reviews posted online<br>- Product review scrapers available<br>- Emergence of "eco-labels" | - Relatively low number of "eco" products available on Amazon<br>- Number of reviews on eco-products inferior to the standard alternatives |
| Data formatting | - Existing (easy-to-use/adapt) codes for automated preprocessing (tokenization, stopwords) | - Filtering out fake positive and negative reviews<br>- Interpretation of irony and sarcasm |
| Model training | - Many open libraries/packages<br>- State-of-the-art pre-trained BERT model for specific fine-tuning | - No library/dictionary trained specifically on sustainable features<br>- Mix (un)supervised ML techniques |
| Model evaluation | - Performance evaluating metrics (easy-to-compute/implement): precision, recall, and F-measure | - Agreed definition of attributes for "sustainable products"<br>- Build an ad-hoc evaluation system |
| Model deployment | - Features valued by customers<br>- Aspect-based sentiment analysis on key and (non-)sustainable features | - Different existing models (e.g. for SA) can provide different results<br>- Meaningful s-design improvement |

## 4. CONCLUSION AND NEXT STEPS

Product development plays a crucial part in creating a sustainable future for all, and the democratization of data is becoming a pillar for new product design and development processes [3]. The methods and tools to develop novel and more sustainable products need to be co-developed together with the latest and state-of-the-art techniques available, including artificial intelligence-based tools such as machine learning for mining and interpreting large amounts of data. As of now, one can argue product (eco-)designers should no longer neglect the processes that lead to new products' perception and choice [9] and should be empowered by the recent advancements in artificial intelligence to design better products [3]. The interplay between emerging topics in product design, such as sustainability, digitalization, and the need for conveying sustainable value through products, has been overlooked in the past.

In this line, the present study investigated the potential of natural language processing techniques to automatically extract sustainable design insights and learning from online product reviews. Leveraging such a data-driven approach for sustainable design through machine learning tools appears to be a timely opportunity: (i) to complement conventional approaches (e.g., focus groups, surveys) aiming at understanding customer preference and needs, and (ii) to develop sound sustainable products, that are valued and used in a sustainable fashion by the customers. In fact, this paper discussed and illustrated how data analytics on online product reviews could be used to discover usage patterns and drive adaptations or redesign of products. Machine learning tools appear to be particularly useful for discovering insights within an extensive collection of data [33], notably in continually changing environments (e.g., new reviews), and can adapt to (and learn from) new scenarios (e.g., new generations of products, new usage patterns).

The present contribution investigates and discusses how machine learning techniques and tools could be deployed to advance the sustainable design and use of products using information from online product reviews. More specifically, the opportunities and challenges of using computer intelligence – including the wealth of Python libraries, packages, and other web-based tools available – in order to get insights from product reviews to design more sustainable products were mapped all along a generic and customizable machine learning process for natural language processing. From that, several analyses useful at the same time for marketers and designers can be imagined, as illustrated in this paper. With the current natural language processing algorithms, the attraction of customers for apparent or alleged sustainable signs can be better understood [34], and product designers can thus further provide eco-friendly customers with what they want to buy [35, 36]. To guarantee the generation of sound sustainable design learnings and leads from online products review, we argue that not only further machine learning models and natural language processing algorithms need to be fine-tuned, but also properly interpreted and deployed by human intelligence.

On this basis, the next steps will consist of: (i) building an integrated and ad hoc machine learning solution (pipeline) to automate this process, by combining and/or training the adequate models and algorithms with state-of-the-art methods in the field of natural language processing, such as BERT [22, 37]; (ii) experimenting the developed solution through case studies comparing standard products against their "eco-friendly" alternatives; (iii) evaluating and fine-tuning the models accordingly; and (iv) scaling and deploying the developed solution, e.g., through an easy-to-use web platform.